\documentclass[letterpaper, 10 pt, conference]{ieeeconf}  

\IEEEoverridecommandlockouts                              

\usepackage{graphics}
\usepackage{multirow}

\usepackage{graphics} 
\usepackage{epsfig} 
\usepackage{times} 
\usepackage{amsmath} 
\usepackage{amssymb}  

\usepackage{array}
\usepackage{booktabs}

\usepackage[hidelinks]{hyperref}
\usepackage{cite}

\usepackage{xcolor,pifont}
\usepackage{color}
\usepackage[nolist]{acronym}
\usepackage{amsmath}
\usepackage{multirow}
\usepackage{microtype}
\usepackage{hyperref}

\newcommand{\Figref}[1]{Figure~\ref{#1}}  
\newcommand{\figref}[1]{Fig.~\ref{#1}}    

\newcommand{\tabref}[1]{Table~\ref{#1}}

\newcommand{\eqnref}[1]{Eq.~\ref{#1}} 
\newcommand{\Real}{\ensuremath{\mathbb R}}  

\DeclareMathOperator*{\MSE}{MSE}
\DeclareMathOperator*{\enc}{\textbf{enc}}
\DeclareMathOperator*{\dec}{\textbf{dec}}

\title{\LARGE \bf
DFM: Deep Fourier Mimic for Expressive Dance Motion Learning
}

\author{Ryo Watanabe$^{1}$$^{,}$$^{2}$, Chenhao Li$^{1}$$^{,}$$^{3}$ and Marco Hutter$^{1}$
\thanks{$^{1}$Ryo Watanabe, Chenhao Li and Marco Hutter are with the Robotic Systems Lab, Department of Mechanical Engineering, ETH Zurich, Switzerland.}
\thanks{$^{2}$Ryo Watanabe is with Sony Group Corporation, Japan}%
\thanks{$^{3}$Chenhao Li is with ETH AI Center, Switzerland}%
}

\begin{document}

\maketitle
\thispagestyle{empty}
\pagestyle{empty}

\begin{abstract}
As entertainment robots gain popularity, the demand for natural and expressive motion, particularly in dancing, continues to rise.
Traditionally, dancing motions have been manually designed by artists, a process that is both labor-intensive and restricted to simple motion playback, lacking the flexibility to incorporate additional tasks such as locomotion or gaze control during dancing.
To overcome these challenges, we introduce \ac{dfm}, a novel method that combines advanced motion representation with \ac{rl} to enable smooth transitions between motions while concurrently managing auxiliary tasks during dance sequences.
While previous frequency domain based motion representations have successfully encoded dance motions into latent parameters, they often impose overly rigid periodic assumptions at the local level, resulting in reduced tracking accuracy and motion expressiveness, which is a critical aspect for entertainment robots.
By relaxing these locally periodic constraints, our approach not only enhances tracking precision but also facilitates smooth transitions between different motions.
Furthermore, the learned \ac{rl} policy that supports simultaneous base activities, such as locomotion and gaze control, allows entertainment robots to engage more dynamically and interactively with users rather than merely replaying static, pre-designed dance routines.
\end{abstract}

\section{INTRODUCTION}
Recent advancements in entertainment robots~\cite{aibo,lovot, disney_learning} have significantly broadened their capabilities, enabling them to perform a wider range of tasks and facilitating more meaningful human-robot interactions~\cite{KANG2020207, companionAIBO2003, robinson2013psychosocial}.
A prominent example is Sony's aibo~\cite{aibo}, a robotic dog designed to engage and captivate its owners.
Within the realm of entertainment robotics, dancing has become one of the most effective ways to attract human attention and express robotic emotions through motion~\cite{dance_robot_god,boston_dynamics_dance,anymal_dance}. 

Typically, the diverse artistic motions performed by these robots are meticulously crafted by designers at entertainment robotics companies~\cite{aibo_dance, disney_learning}.
This process is inherently time-consuming, requiring motion designers to carefully create and fine-tune various dance motions, most of which are handcrafted.
Despite the presence of common motion primitives within these dance routines, designers still need to create complete dance motions for each instance, which demands significant time and effort.
Additionally, these artistic motions are often replayed with less emphasis on interacting dynamically with humans.

The field of computer animation has proposed various methods to streamline the motion creation process.
A promising approach involves the use of learning-based methods to generate diverse motions.
Recent advancements in this area have leveraged \ac{gan}~\cite{goodfellow2020generative} to address this issue~\cite{amp_org,amp_quadruped_robot,li2023learning,peng2022ase,li2023versatile,luo2023perpetual,tessler2023calm}.
However, these methods still face limitations in handling a wide variety of demonstration motion datasets due to mode collapse.
An alternative approach focuses on motion representation.
Several studies~\cite{periodic_autoencoder, ai_choreographer, transflower, fld} have explored methods to automatically align and represent periodic dance motions within a structured latent space. 
For instance, the \ac{fld} approach~\cite{fld}, which integrates \ac{pae}~\cite{periodic_autoencoder} and \ac{rl}, addresses some of these challenges by conditioning desired motion trajectories during policy inference.
The structured latent space in \ac{fld} allows for the creation of diverse motions through smooth transitions between motion primitives, thereby reducing the time-intensive process of motion creation for designers.
However, the reliance on strong periodicity assumptions in \ac{fld}, even at the local level, leads to overly smoothed motions and a loss of expressiveness, which is particularly problematic for entertainment robots where dynamic and expressive motions are essential.

\begin{figure}[!t]
    \centering
    \includegraphics[width=\linewidth]{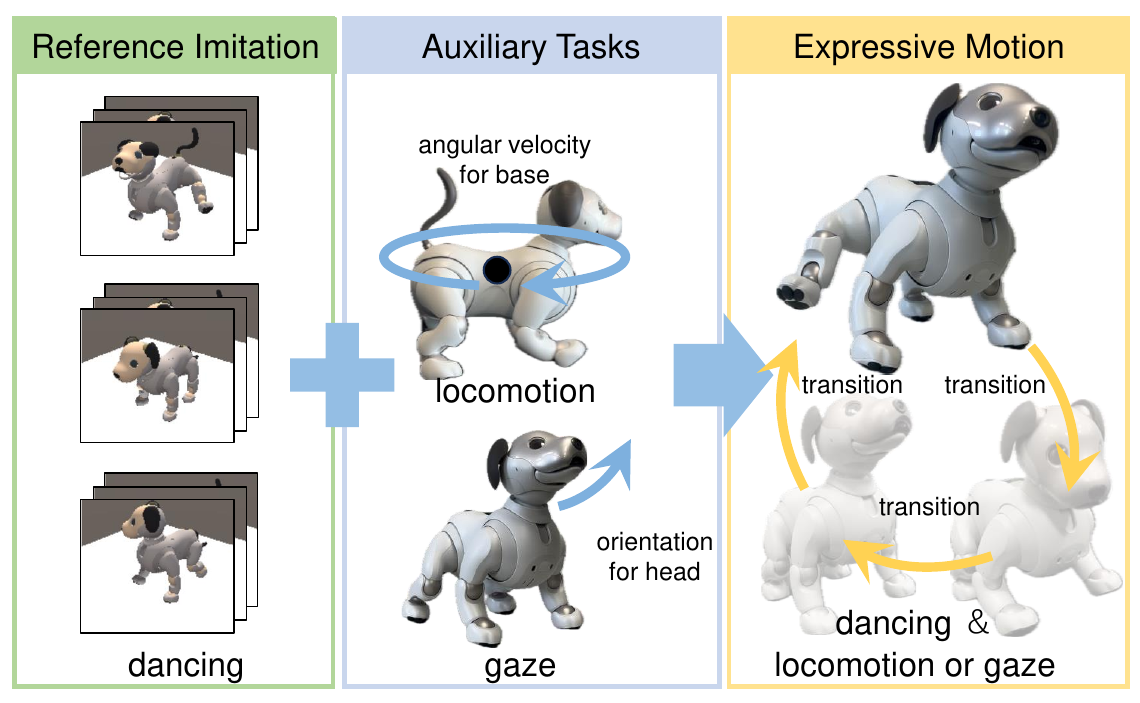}
    \caption{Deep Fourier Mimic (DFM) allows entertainment robots such as aibo to seamlessly combine artistic motions crafted by designers with auxiliary tasks like locomotion or gaze towards a human face, resulting in expressive motion that can smoothly transition between different movements at arbitrary timings. Project webpage: \url{https://sony.github.io/DFM/}}
    \label{fig:dfm_concept}
    \vspace{-3ex}
\end{figure}

To extend the capabilities of \ac{fld} to general, less periodic motions such as dancing, we propose \ac{dfm} as shown in \figref{fig:dfm_concept}, a method that trains policies using fresh encodings of the most recent motion segments, thereby relaxing the strong local periodic assumptions to preserve expressiveness and variation in the generated motions.
Additionally, to increase human interaction during stylized dancing, we introduce supplementary tasks such as locomotion and head orientation tracking called gaze.
For instance, maintaining eye contact by gazing at a human face, which mimics real dog behavior, can create a more interactive and engaging experience during the dance with a whole-body approach.

Our primary contributions are as follows:
\begin{itemize}
\item Development of a method for learning from demonstrations that achieves high tracking accuracy of reference motions while preserving details and ensuring smooth transitions between various dance sequences for entertainment robots.
\item Demonstration and validation in hardware experiments of multi-task capabilities, including auxiliary gaze and locomotion control during dance motion, coordinated in a whole-body manner that integrates both leg and head movements.
\end{itemize}

\begin{figure*}[!t]
    \centering
    \includegraphics[width=\linewidth]{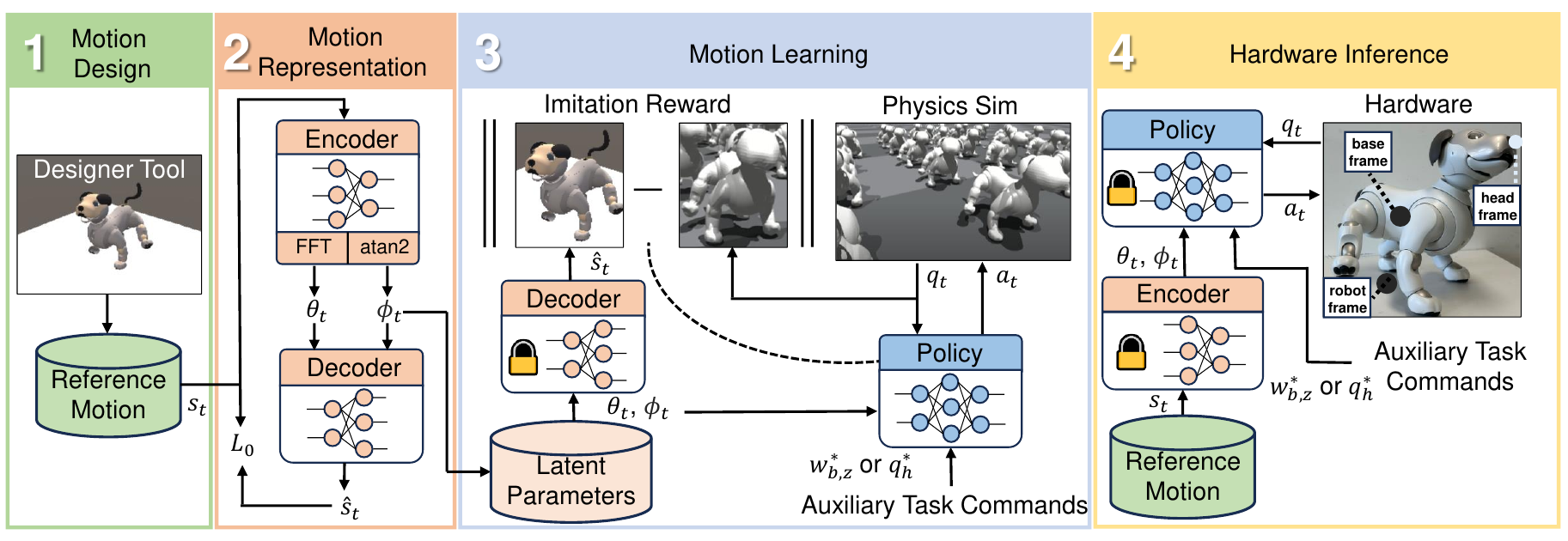}
    \caption{The expressive dance motion learning system is composed of four key components: motion design, motion representation, motion learning, and hardware inference. In the motion design phase, artists create motion references using specialized design software. The representation of these diverse motions is then learned using a \ac{pae}. Reinforcement learning (\ac{rl}) is employed to enable the robot to perform auxiliary tasks, such as walking and head orientation control, while accurately tracking the designed dance references. During inference, the learned policy is deployed on the actual hardware, allowing for real-time execution of dance motions and dynamic and interactive motions by tracking the auxiliary task commands.}
    \label{fig:system_overview}
    \vspace{-2ex}
\end{figure*}

\section{RELATED WORK}
\subsection{Robotic Dance}
The development of dancing for entertainment robots has seen significant progress~\cite{dance_robot_god}.
Nakaoka et al. employed a motion capture system to teach the HRP-2~\cite{HRP2} a traditional Japanese folk dance~\cite{dancing_humanoid}.
While this approach showcased impressive dancing capabilities, the sequences and timings of all movements were meticulously pre-programmed to align with a specific music.
For quadruped robots, a notable example is a dancing robot capable of detecting tempo and adjusting its motions in real-time~\cite{anymal_dance}.
Similarly, certain entertainment robots, such as Sony's aibo, incorporate the ability to adjust their dancing tempo~\cite{aibo_dance}.
However, these systems rely mostly on replaying predefined motion sequences and lack the capability to perform additional task operations during motion execution, which is a key focus of this paper.

\subsection{Learning from Demonstrations}
Reinforcement learning (\ac{rl}) has been extensively utilized to generate robust locomotion policies for quadrupedal robots~\cite{anymal_terrain, anymal_perceptive, Choi2023-cf}.
Despite these advances, achieving stylized motions that appear natural to human observers remains challenging.
To address the need for natural motion learning, Peng et al.\cite{deepmimic} proposed a method that enhances \ac{rl} by incorporating rewards to imitate reference trajectories within a physics-based simulator.
Recently, Ruben et al.\cite{disney_learning} applied this approach to real robots.
However, these methods are constrained by their reliance on a single trajectory, making it difficult to transition between reference motions without intricate motion design.

To handle transitions between multiple reference motions, various methods based on \ac{gan} have been developed~\cite{amp_org, amp_quadruped_robot}.
Nevertheless, these approaches are often plagued by mode collapse, a phenomenon resulting from insufficient motion distinction without intrinsic diversity rewards.
Although recent works~\cite{peng2022ase, li2023versatile, luo2023perpetual, tessler2023calm} have made efforts to mitigate mode collapse, they still struggle to handle the switching of specific motions at arbitrary timings, a critical requirement for entertainment robots.

In contrast, self-supervised representation methods using physics simulators have emerged as another promising research direction~\cite{motion_representation}.
While some previous studies have specifically focused on learning dancing motions~\cite{ai_choreographer, transflower}, the \ac{pae}~\cite{periodic_autoencoder} has proven effective in producing smoother and more stable movements for periodic motions, such as locomotion and dancing, with a special treatment of the spatial-temporal relationships inherent in the motions.
The \ac{fld} approach\cite{fld} further extends \ac{pae} by introducing latent dynamics with a quasi-constant representation of motions that greatly reduces the parameterization effort in motion representation and facilitates downstream learning with \ac{rl}.
These methods automatically align and represent motions within a structured latent space, allowing for conditioning desired motion trajectories during policy inference without suffering from mode collapse.
However, both the representation and policy learning in \ac{fld} impose strong periodic assumptions, which can lead to over-smoothed motions that lack expressiveness, a significant drawback for entertainment robots that aim to replicate the nuanced motions of real animals.

\section{METHOD}
As illustrated in \figref{fig:system_overview}, \ac{dfm} training pipeline for expressive dance motion learning system comprises four key components: motion design, motion representation, \ac{rl} training, and hardware inference.

\subsection{Motion Design}
Reference motions, including joint positions and velocities for each joint, are created using specialized design tools.
Following practices in the animation industry, our motion data are crafted by artistic designers.
We select 34 distinct dance motions, all created by motion designers at Sony, as showcased in the supplementary video.
Each motion is augmented to include five frequency variations — 0.5, 0.75, 1.0, 1.25, and 1.5 — by sampling and interpolating the motion trajectories.
This process results in a total of 170 clips with a duration of 6 seconds each.

\subsection{Motion Representation}
Motions are commonly described as long-horizon trajectories in high-dimensional state space.
However, directly associating motions with raw trajectory instances yields highly inefficient representations and poor generalization, failing to develop a policy that naturally transits in between.
We denote trajectory segments of length $H$ in $d$-dimensional state space preceding time step $t$ by $\mathbf{s}_t = (s_{t-H+1}, \dots, s_{t}) \in \Real^{d \times H}$.
In the motion representation stage, we utilize \ac{fld}~\cite{fld} to encode motion trajectories into latent parameters, which consist of $\theta_t = (f_t, a_t, b_t)$ and $\phi_t$, where $f_t$, $a_t$, $b_t$ and $\phi_t$ denote latent frequency, amplitude, offset and phase, respectively.
The \ac{pae} structure featured by \ac{fld} employs a set of encoder and decoder composed of 1D convolutional layers through time~\cite{periodic_autoencoder}.
While $\theta_t$ are computed with a differentiable real \ac{fft} layer, the latent phase $\phi_t$ is determined using a linear layer followed by atan2 applied on 2D phase shifts on each channel as shown in \figref{fig:system_overview}.
With the characteristic convolution and \ac{fft} layers, the \ac{fld} encoder decomposes the input motions into a latent phase parameter $\phi_t$ representing the low-level local time index and a latent frequency domain parameter $\theta_t$ representing high-level global features, respectively.
In the original paper, \ac{fld} formulates latent dynamics with latent frequency $f_t$ and time increment $\Delta t$ is thus described as
\begin{equation}
\begin{split}
    \theta_t = \theta_{t-1}, \quad \phi_t = \phi_{t-1} + f_{t-1} \Delta t.
    \label{eqn:latent_dynamics}
\end{split}
\end{equation}
The encoder and decoder are described with $c$ latent channels, 
\begin{equation}
    \phi_t, \theta_t = \enc(\mathbf{s}_t), \quad \hat{\mathbf{s}}'_{t+i} = \dec(\phi_t + i f_t \Delta t, \theta_t),
    \label{eqn:periodic_assumption}
\end{equation}
where $\phi_t \in \Real^c$ and $\theta_t \in \Real^{3c}$.

Fourier Latent Dynamics (\ac{fld}) introduces a forward prediction mechanism in the latent representation space, where it reconstructs future states with phase propagation while maintaining $\theta_t$.
The proceeding motion segment $\mathbf{s}_{t+i} = (s_{t-H+1+i}, \dots, s_{t+i})$ is approximated with the prediction $\hat{\mathbf{s}}'_{t+i}$ decoded from $i$-step forward propagation using the latent dynamics from time step $t$.
The latent dynamics in \eqnref{eqn:latent_dynamics} assumes locally constant latent parameterizations and propagates latent states by advancing $i$ local phase increments.
Therefore, the total loss for training \ac{fld} with $N$-step forward prediction by propagating the latent dynamics is written as
\begin{equation}
\begin{split}
    L_{FLD}^N = \sum_{i=0}^N \MSE(\hat{\mathbf{s}}'_{t+i}, \mathbf{s}_{t+i}).
    \label{eqn:total_loss}
\end{split}
\end{equation}

The choice of the number of forward prediction steps, $N$, plays a crucial role in balancing the trade-off between the accuracy of local reconstructions and the global coherence of the latent parameterizations.
In the original \ac{fld} approach, a larger value of $N$, such as $N = 100$ at $\Delta t = 0.01$, is typically selected to emphasize the global consistency of the latent parameters within periodic motions to facilitate downstream \ac{rl} training.
However, this can result in reduced reconstruction accuracy of general, less periodic reference trajectories, which is particularly critical for entertainment robots where precise motion tracking is essential.
To address this, \ac{dfm} sets $N=0$, prioritizing local reconstruction accuracy to enhance the tracking performance while maintaining a fresh local motion encoding during policy training.
The specific hyperparameters used for the motion representation are detailed in \tabref{table:fld_param}. 

\begin{table}[t]
\caption{Motion Representation Hyper Parameters}
\label{table:fld_param}
\begin{center}
\begin{tabular}{cccc}
\toprule
\textbf{Entry} & \textbf{Symbol} & \textbf{Value} \\
\midrule
step time seconds & $\Delta t$ & 0.01 \\
mini batch size & ${-}$ & 50 \\
learning rate & ${-}$ & 0.0001 \\
weight decay & ${-}$ & 0.0005 \\
max iterations & ${-}$ & 5000 \\
latent channel number & ${c}$ & 8 \\
periodic trajectory segment  & ${H}$ & 100 \\
forward prediction step & ${N}$ & 0  \\
\bottomrule
\end{tabular}
\end{center}
\end{table}

\begin{table}
\caption{Policy Observation}
\label{table:observation}
\begin{center}
\begin{tabular}{cccc}
\toprule
\textbf{Entry} & \textbf{Symbol} & \textbf{Noise Level} & \textbf{Dimensions} \\
\midrule
joint positions & ${q}$ & 0.01 & 14 \\
joint velocities & ${\dot{q}}$ & 1.5 & 14  \\
last action & ${a}^*$ & 0.0 & 14 \\
foot contact state & ${f_c}$ & 0.0 & 4 \\
gravity orientation & ${g}$ & 0.05 & 3 \\
latent phase & ${\sin{\phi}}$ & 0.0 & 8 \\
latent phase & ${\cos{\phi}}$ & 0.0 & 8 \\
latent frequency & ${f}$ & 0.0 & 8 \\
latent amplitude & ${a}$ & 0.0 & 8 \\
latent offset & ${b}$ & 0.0  & 8\\
\bottomrule
\end{tabular}
\end{center}
\end{table}

\begin{table*}[t]
\caption{Reward functions and scales at imitation and curriculum phase for locomotion and gaze}
\label{table:reward}
\begin{center}
\begin{tabular}{lllccc}
\toprule
\textbf{Category} & \textbf{Reward} & \textbf{Definition} & \textbf{Dance} & \textbf{Locomotion} & \textbf{Gaze}\\
&&& \textbf{imitation} & \textbf{curriculum} & \textbf{curriculum} \\
\midrule
\multirow{1}{*}{\textbf{Imitation}} & joint position imitation & $\exp(\|{q}^* - {q}\|^2 )$ & $1.0$ & $1.0$ & $1.0$ \\
\midrule
\multirow{2}{*}{\textbf{Task}} 
 & base angular velocity tracking & $\exp(-\frac{1}{0.06}\|{w}_{b,z}^* - {w}_{b,z}\|^2)$ & $0.0$ & $1.0$ & - \\
 & end-effector orientation tracking  & $\exp(4\|{q}_{h}^* - {q}_{h}\|^2 )$ & $0.0$ & - & $0.7$ \\
\midrule
\multirow{6}{*}{\textbf{Regularization}} & joint torque & $\|{\tau}\|^2$ & $-0.001$ & $-0.001$ & $-0.001$  \\
 & joint acceleration & $\|{\ddot{q}}\|^2$ & $-2e^{-7}$ & $-2e^{-7}$ & $-2e^{-7}$ \\
 & joint target difference & $\|{a}^*_{t-1} - {a}^*_{t}\|^2$ & $-0.01$ & $-0.01$ & $-0.01$ \\
 & self-collisions & $n_{c}$ & $-10.0$ & $-10.0$ & $-10.0$ \\
 & foot slippage & $\|{v}_{f,xy}\|^2$ & $0.0$ & $-0.15$ & - \\
 & foot air time & $\sum_i ({t}_{f,air} - $0.2$)$ & $0.0$ & $2.0$ & - \\
\bottomrule
\end{tabular}
\end{center}
\end{table*}

\subsection{Motion Learning}
Following the motion representation stage, \ac{dfm} employs \ac{rl} to track the encoded motions while managing additional tasks such as locomotion and gaze control during stylized dancing, as depicted in \figref{fig:system_overview}.
The training network architecture consists of three fully connected layers, each with 256 hidden units and ELU activation functions, and utilizes the \ac{ppo}~\cite{ppo}.
Both the actor and critic networks are implemented as \ac{mlp}, adhering to the architecture and hyperparameters established in prior work~\cite{fld}.
The simulation and control loop frequencies are set to 400 Hz and 100 Hz, respectively, using the Isaac Gym framework~\cite{Makoviychuk2021-th}.

The aibo features 14 \ac{dof}, including 12 \ac{dof} for its legs and 2 \ac{dof} for head pitch and yaw movements, represented by the action vector ${a}^*$.
Policy observations and noise levels are detailed in \tabref{table:observation}.
Given that aibo is a consumer-grade robot, ${f_c}$ values are obtained from binary switch contact sensors.
The latent parameters with 8 channels are used as observations.
Unlike \ac{fld}, which assumes time-invariant latent parameters $\theta_t$ for each episode, \ac{dfm} updates $\theta_t$ and $\phi_t$ dynamically during motion representation to capture the nuanced details of the artistic reference motions.
This allows for the expression of critical changes in frequency, amplitude, and offset, preserving details in expressive movements.

In contrast to \ac{fld}, which limits motion representation and learning only to locomotion tasks, the extended capabilities of \ac{dfm} are demonstrated with distinct tasks assigned to each stage.
Specifically during the \ac{rl} phase, the policy is conditioned to track a target motion sequence using the learned representations, while locomotion and gaze are trained with auxiliary task objectives.
The task command for locomotion is the angular velocity of the base axis, while for gaze is the pitch and yaw angles of the head axis in the robot frame, as illustrated in \figref{fig:system_overview}.

The reward functions and their corresponding scales are outlined in \tabref{table:reward}.
The variables ${q}$, ${\tau}$, and ${n}_{c}$ represent joint positions, joint torques, and the number of collisions, respectively.
${q}_{h}$ denotes the yaw and pitch angles of the head axis, with ${q}_{h}^*$ representing their commanded values.
Similarly, ${{w}_{b}}$ and ${{w}_{b}}^*$ denote the measured and desired angular velocities of the base axis, respectively.
Imitation rewards are designed to ensure the policy closely mimics the reference dancing motion.
Joint imitation rewards are computed as the difference between the reconstructed joint positions and the current joint position in the physics simulator.
Once the agent achieves a joint position imitation reward higher than 0.9, task rewards for locomotion or gaze are introduced.
Among the regularization rewards, joint target difference, joint torque, and joint acceleration rewards suppress jerky motions, while self-collision rewards prevent collisions between the robot’s parts.
Foot velocities ${{v}_{f}}$ for foot slippage and airtime for foot parts ${t}_{f,air}$ are monitored to shape walking during the locomotion curriculum phase.

\subsection{Hardware Inference}
As depicted in \figref{fig:system_overview}, the reference motion, encoder, and policy, developed during the motion design, motion representation, and motion learning stages, respectively, are deployed on the robot.
The \ac{rl} policies for locomotion and gaze are implemented as separate entities, with task commands being switched accordingly for each policy.
Both control and sensor reading operations are performed at a frequency of 100 Hz.
The gravity orientation is calculated by fusing acceleration and gyroscope data from the IMU.

To ensure smooth transitions between different motions at arbitrary times, we interpolate the latent parameters during the transition period using \eqnref{eqn:natural_transient}:
\begin{equation}
\begin{split}
    \theta_{A\rightarrow B} = \alpha \theta_A + (1 - \alpha) \theta_B, \\ 
    \phi_{A\rightarrow B} = \alpha \phi_A + (1 - \alpha) \phi_B,
\end{split}
    \label{eqn:natural_transient}
\end{equation}
where $\theta_A$ and $\phi_A$ represent the final latent parameters of motion A, and $\theta_B$ and $\phi_B$ correspond to the initial latent parameters of motion B. The interpolation factor $\alpha$ is smoothly varied from $0$ to $1$ over a designated transition period, which in our implementation is set to 0.5 seconds.
\section{Experiments and Results}
We evaluate the performance of \ac{dfm} in terms of tracking accuracy relative to reference motions and the naturalness of transitions between motions. 
Additionally, we demonstrate the multitasking capabilities of our approach through locomotion and gaze control during stylized dancing.

\subsection{Tracking Accuracy}
\begin{figure}[t]
    \vspace{1ex}
    \centering
    \includegraphics[trim={0 0 0 0}, width=\linewidth]{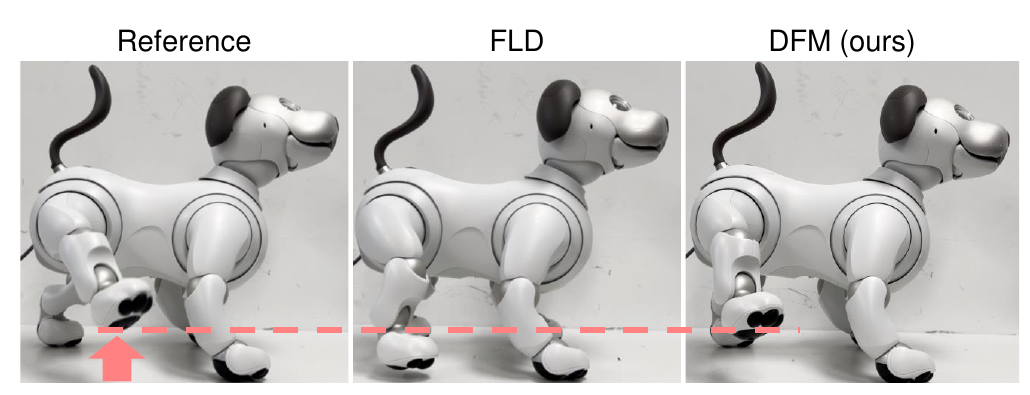}
    \caption{Height reached by rear right leg. Left, middle and right depict reference motion, \ac{fld} and \ac{dfm} motions respectively. The red dash line illustrates the height of the right rear leg at reference motion.}
    \label{fig:moveup_rear_leg}
    \vspace{-1ex}
\end{figure}

\begin{figure}[t]
    \centering
    \includegraphics[trim={0 2ex 0 0}, width=\linewidth]{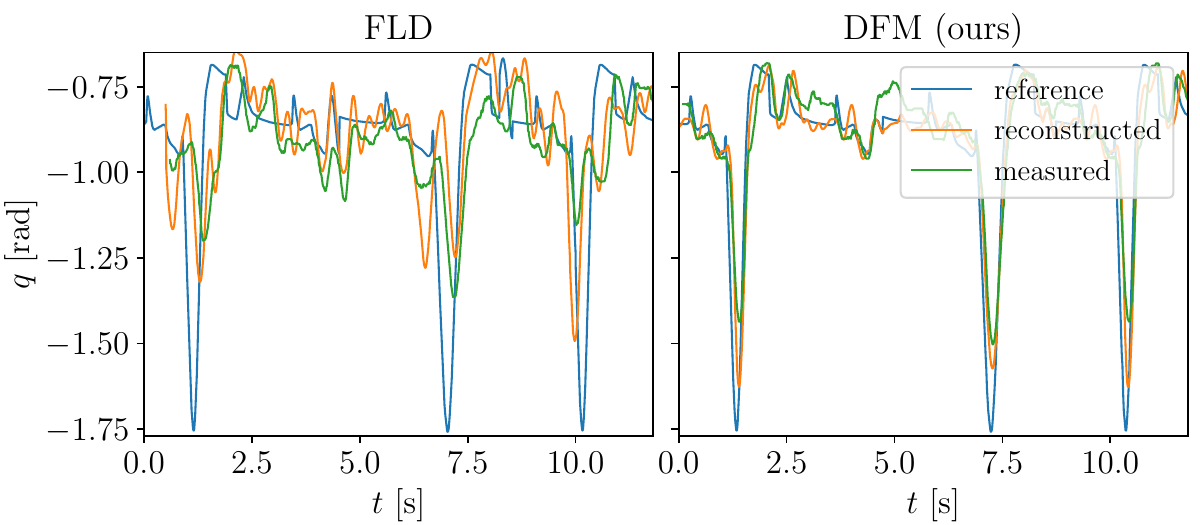}
    \caption{Comparison of tracking accuracy for the \ac{fld} and \ac{dfm}. Blue: reference motion created by the motion designer. Orange: reconstructed motions from motion representation parts by conditioning the reference motion. Green: joint encoder reading activated by the \ac{rl} policy.}
    \label{fig:joint_angle_tracking_acc}
    \vspace{-3ex}
\end{figure}

\begin{figure}[!t]
    \vspace{1ex}
    \centering
    \includegraphics[trim={0 0 0 0}, width=\linewidth]{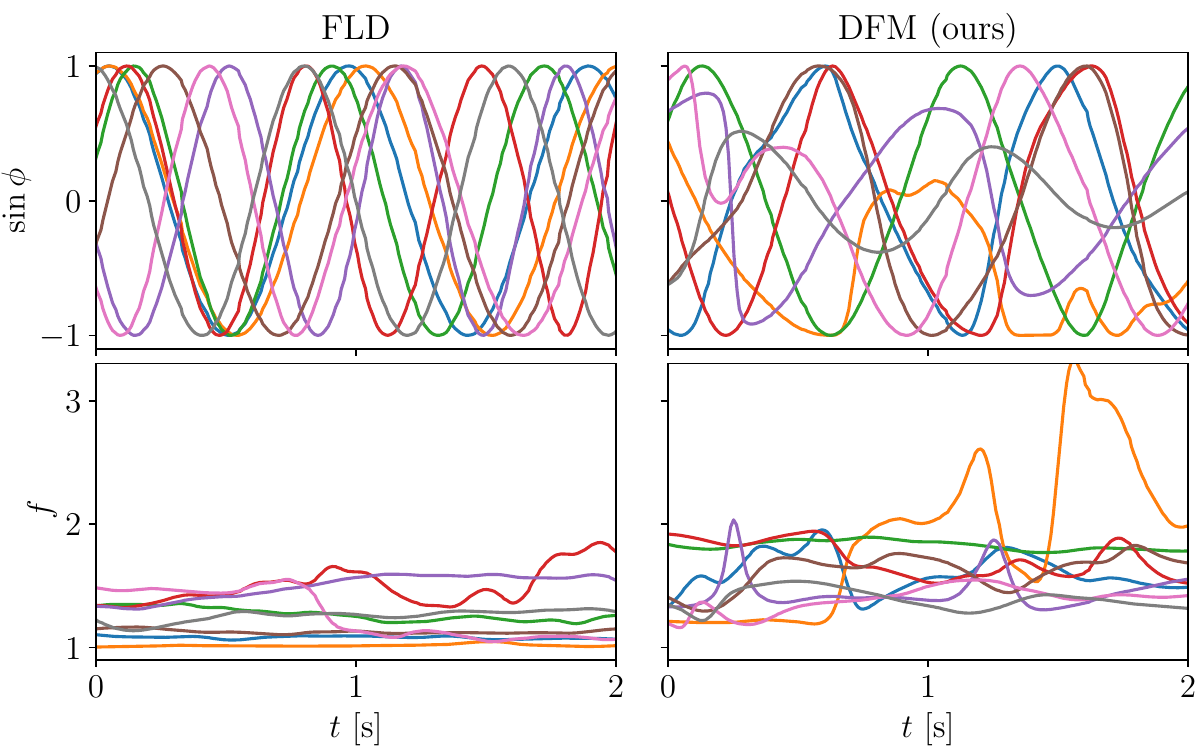}
    \caption{Comparison of 8 channel latent parameters for \ac{fld} at the left and \ac{dfm}  at the right side by conditioning the same dancing motion as \figref{fig:moveup_rear_leg}. The upper and bottom of plots are $\sin{\phi}$ and frequency for each.}
    \label{fig:latent_parameters}
\end{figure}

To assess tracking accuracy, we calculate the difference between the reference motion and the observed joint positions on the robot hardware.
For this evaluation, we select a dancing motion that involved lifting the rear legs.
\figref{fig:moveup_rear_leg} illustrates the height of the rear right leg during this stylized dance.
Although \ac{dfm} does not perfectly replicate the height of the reference motion, it significantly outperforms the \ac{fld} baseline.
To provide a more quantitative comparison between the baseline and our method, we analyze three types of joint angles as shown in \figref{fig:joint_angle_tracking_acc}.
The results indicate that the motion reconstructed by \ac{fld} is overly smooth due to its strong enforcement of quasi-constant parameterization and periodicity assumption with $N = 100$.
In contrast, \ac{dfm} achieves a more accurate reconstruction with $N = 0$, preserving intricate details that may not follow periodic patterns.
When examining the joint encoder data measured from the robot, \ac{fld} again shows excessive smoothing, which we attribute to the overly strong periodic assumptions applied to the local time during \ac{rl} training, as described in \eqnref{eqn:periodic_assumption}.
\Figref{fig:latent_parameters} presents the $\sin{\phi}$ and frequency values derived from the latent parameters across eight channels during the same dancing motion.
On the left side, \ac{fld} shows that all channels of $\sin{\phi}$ are periodic, with little change in frequency.
In contrast, \ac{dfm} demonstrates variability in some channels of $\sin{\phi}$ and frequency during the upward movement of the rear leg, retaining non-periodic features that characterize the dance motions.
Finally, \tabref{table:tracking_accuracy} reports the mean absolute tracking error (MAE) across all joints for all 170 evaluated motions using the real aibo hardware.
Additionally, we test \ac{dfm} using the MIT Humanoid environment~\cite{chignoli2021humanoid} in Isaac Gym.
Our method consistently demonstrates superior tracking accuracy in both robot environments compared to \ac{fld}.

\begin{table}[!t]
\caption{Mean Absolute Tracking Accuracy}
\label{table:tracking_accuracy}
\begin{center}
\begin{tabular}{llcc}
\toprule
\textbf{Robot} & \textbf{reference motion} & \textbf{FLD} & \textbf{DFM (ours)} \\
\midrule
aibo & dance & $0.132$ $\rm{rad}$ & $0.094$ 
 $\rm{rad}$ \\
MIT humanoid  & locomotion &  $0.125$ $\rm{rad}$  & $0.103$ $\rm{rad}$   \\
\bottomrule
\end{tabular}
\end{center}
\vspace{-5ex}
\end{table}

\subsection{Natural Transition}
The motion representation employed by \ac{dfm} enables continuous frequency interpolation and smooth transitions between different dancing motions.

\begin{figure}[!t]
    \centering
    \includegraphics[trim={0 0 0 0}, width=\linewidth]{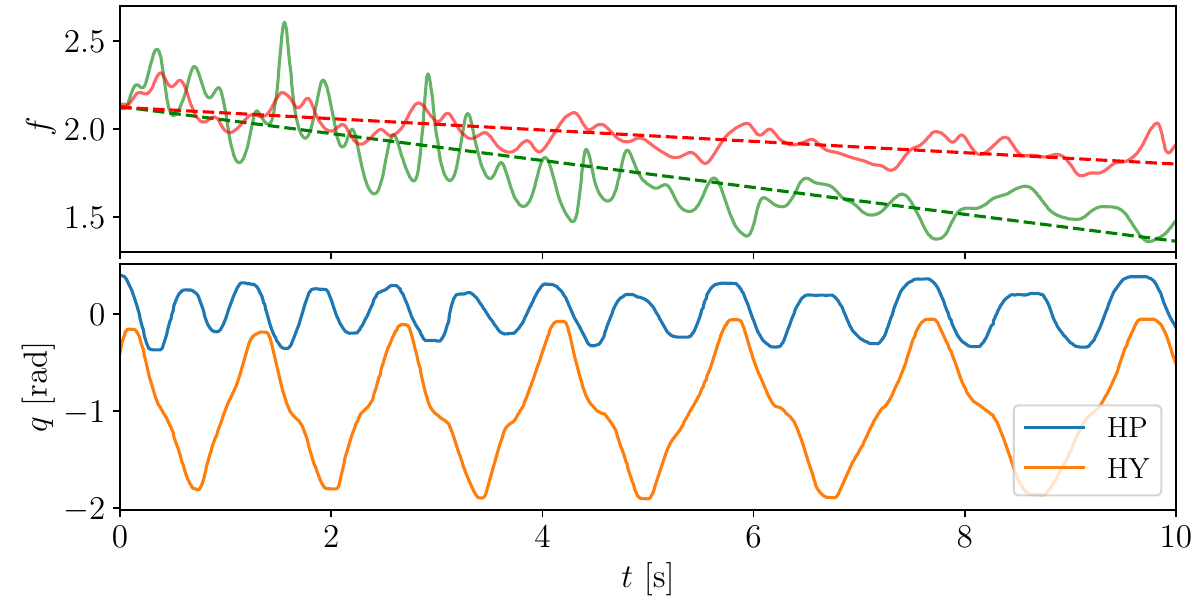}
    \caption{Frequency modulation during head-moving dance. The upper plot displays the frequency of two representative latent channels out of eight. Solid and dashed curves represent raw and linearly interpolated data, respectively. The bottom plot shows head pitch (HP) and the head yaw (HY) joint angles.}
    \label{fig:frequency_interpolation}
    \vspace{-1ex}
\end{figure}

\Figref{fig:frequency_interpolation} shows the estimated latent frequency parameters conditioned on the reference motion, which primarily involves head movements transitioning from higher to lower dancing frequencies.
While most frequency channels remain relatively constant, channels 3 and 4 exhibit gradual changes as shown in \figref{fig:frequency_interpolation}.
The linear interpolation of frequencies in these channels adjusts in response to the changing frequency of the reference dancing motion.
Even though the training dataset consists of discrete frequency types, the motion representation allows for continuous frequency interpolation.
This capability results in smooth, periodic changes in joint positions without abrupt movements, even for previously unseen datasets as shown at the bottom plot in \figref{fig:frequency_interpolation}.

\begin{figure}[!t]
    \centering
    \includegraphics[trim={0 0 0 0}, width=\linewidth]{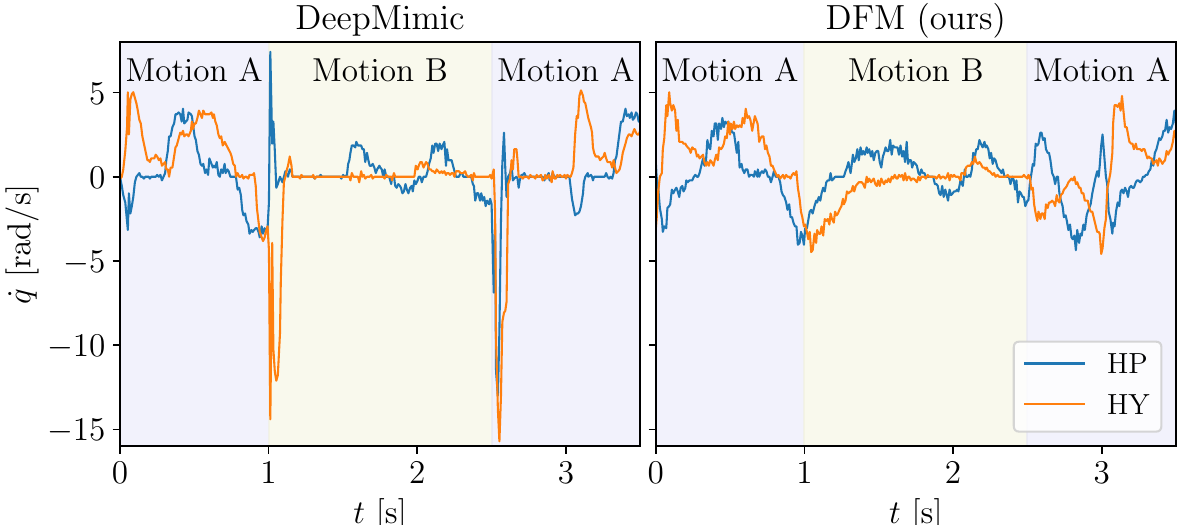}
    \caption{Transition between different dance types. The background color indicates the dance motion type. The left and right plots demonstrate hard switches between Dance A and Dance B, with DeepMimic and \ac{dfm}, respectively. Angular velocities at the head pitch (HP) and the head yaw (HY) are shown.}
    \label{fig:natural_transient}
    \vspace{-3ex}
\end{figure}

\Figref{fig:natural_transient} illustrates the joint angular velocity at the head pitch and yaw during the transition from motion A to motion B, which primarily involves the head pitch and yaw actuators, as shown in the supplementary video.
During the transition times at 1 and 2.5 seconds, joint positions experience abrupt changes with switches between reference motions.
We compare the transition performance of \ac{dfm} with DeepMimic~\cite{deepmimic}, a well-known learning from demonstration approach that yields high tracking performance on single trajectories but lacks capabilities to deal with multiple motions.
Jerky transitions are observed in this case if the reference dataset and its representation are not carefully crafted.
In contrast, \ac{dfm} achieves smooth transitions without abrupt movements by interpolating in the latent space using \eqnref{eqn:natural_transient}.

\subsection{Multi-task Demonstration}
We evaluate the multitasking capability of \ac{dfm} with auxiliary tasks, including locomotion and gaze, respectively.

\begin{figure}[!t]
    \centering
    \includegraphics[trim={0 0 0 0}, width=\linewidth]{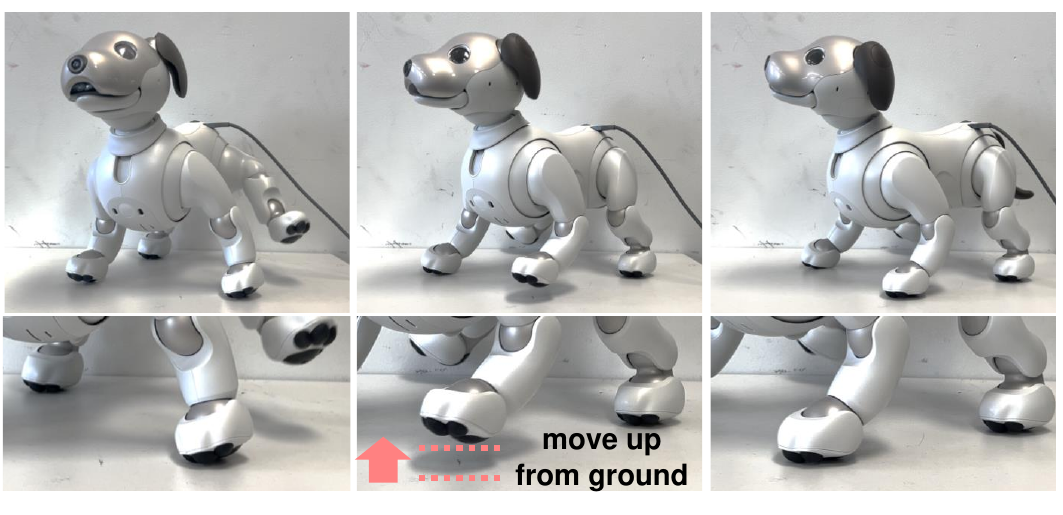}
    \caption{Locomotion during dance. The reference dance motion alternates lifting the rear legs while keeping the forelegs stationary. Applying an angular velocity command results in locomotion by lifting the left foreleg.}
    \label{fig:dancing_locomotion}
\end{figure}

\Figref{fig:dancing_locomotion} illustrates the locomotion policy during dancing, where an angular velocity command is used to facilitate in-place rotation.
In the reference motion, only the rear legs move alternately, while the forelegs remain stationary.
After training the policy with the reward structure defined in the locomotion curriculum (\tabref{table:reward}), aibo learns to rotate in response to the angular velocity command in the base frame.
To allow this rotation without hindering the movement of the rear legs, the right foreleg is lifted, enabling the execution of the stylized dancing, as shown in the supplementary video.

\begin{figure}[!t]
    \vspace{-2ex}
    \centering
    \includegraphics[trim={0 0 0 0}, width=\linewidth]{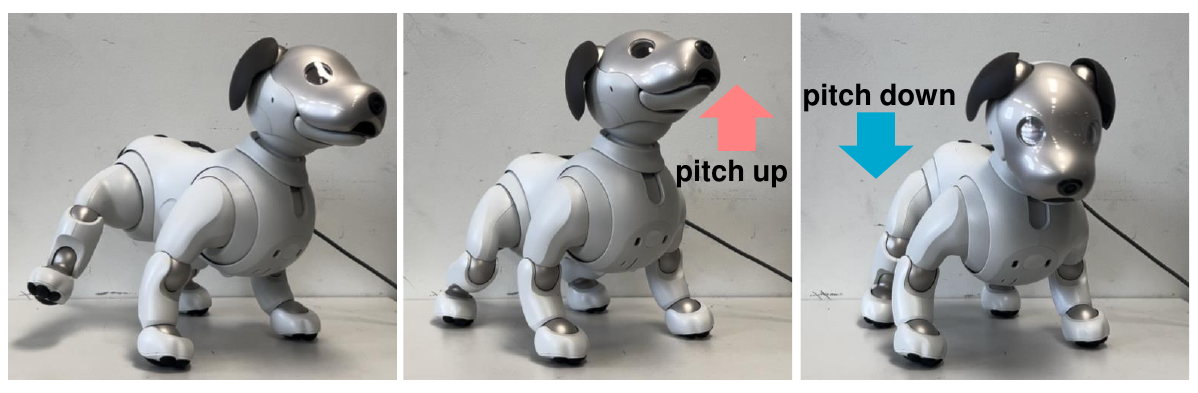}
    \caption{Gaze during dance. The reference motion is the same as in \figref{fig:dancing_locomotion}. The images depict commanded pitch angles of $0.0$, $0.3$ $\rm{rad}$, and $-0.5$ $\rm{rad}$, respectively. The command of the yaw angle is held at zero.}
    \label{fig:dancing_gaze}
\end{figure}

\begin{figure}[!t]
    \centering
    \includegraphics[trim={0 0 0 0}, width=\linewidth]{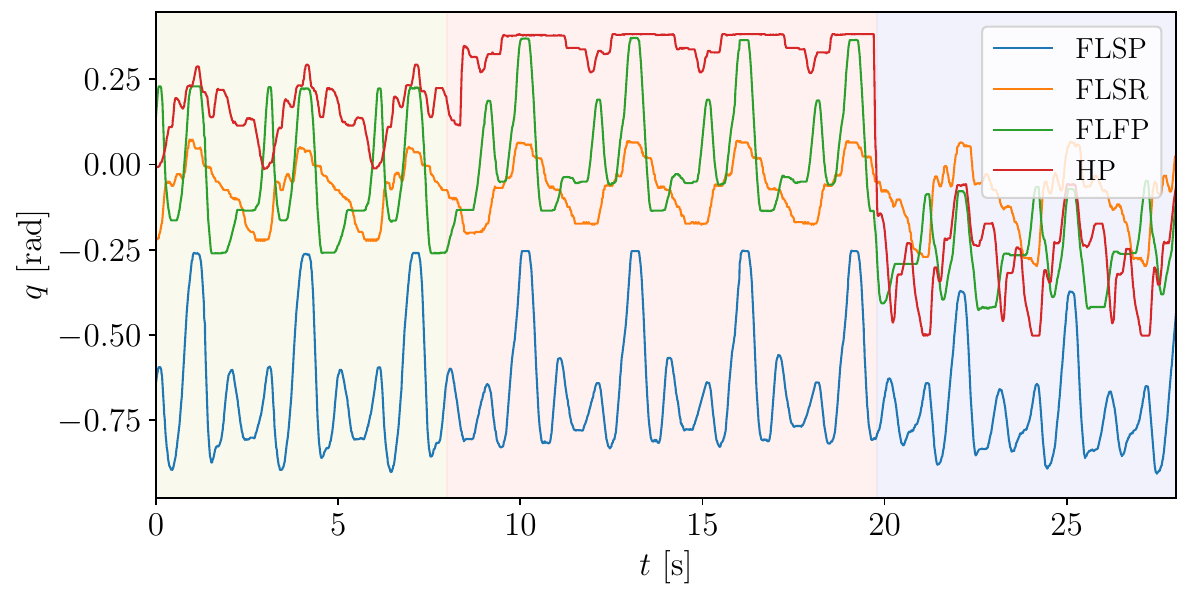}
    \caption{Joint readings for fore left shoulder pitch (FLSP), fore left shoulder roll (FLSR), fore left foot pitch (FLFP), and head pitch (HP) during dancing gaze are shown. The background color of the plot indicates command for pitch angle at the head frame (yellow: $0.0$, red: $0.3$ $\rm{rad}$, blue: $-0.5$ $\rm{rad}$).}
    \label{fig:dancing_gaze_plot}
    \vspace{-3ex}
\end{figure}

Similarly, a policy for auxiliary gaze control is trained using the reward scale from the gaze curriculum in \tabref{table:reward}.
This policy enables aibo to adjust its head orientation in response to pitch and yaw commands during dancing, as demonstrated in \figref{fig:dancing_gaze}.
The supplementary video shows that the dance sequence continues smoothly while the robot adjusts its pitch and yaw angles.
aibo utilizes its head and legs to track the commanded pitch and yaw angles, as illustrated in \figref{fig:dancing_gaze_plot}.
For instance, when a pitch of $0.3$ $\rm{rad}$ and a yaw of $0.0$ $\rm{rad}$ are commanded, both the directions of head are moved up with legs.
In contrast, a $-0.5$ $\rm{rad}$ pitch command prompts the head and legs to move in opposite directions.

\section{CONCLUSION}
We introduce \ac{dfm}, a novel approach that combines motion representation with \ac{rl} to achieve high tracking accuracy, smooth transitions, and the capability to perform additional tasks during dancing.
By relaxing the strong periodic assumptions presented in previous works, our method demonstrates superior motion expressiveness compared to the baselines.
Our motion representation technique enables continuous frequency changes for unseen reference datasets and facilitates smooth transitions between different motion types through latent space interpolation.
Additionally, \ac{dfm} demonstrates extended multi-task capabilities, such as locomotion and gaze control during dancing, leading to more interactive motions beyond simple motion replay.
These findings have significant implications for advancing research in human-robot interaction for expressive entertainment robots.



\begin{acronym}
\acro{gan}[GANs]{Generative Adversarial Networks}
\acro{rl}[RL]{Reinforcement Learning}
\acro{pae}[PAE]{Periodic Autoencoder}
\acro{fld}[FLD]{Fourier Latent Dynamics}
\acro{ppo}[PPO]{Proximal Policy Optimization}
\acro{fft}[FFT]{Fast Fourier Transform}
\acro{pca}[PCA]{Principal Component Analysis}
\acro{dfm}[DFM]{Deep Fourier Mimic}
\acro{dof}[DoF]{Degrees of Freedom}
\acro{mlp}[MLPs]{Multi-Layer Perceptrons}
\end{acronym}

\section*{ACKNOWLEDGMENT}
The authors would like to thank Hiroyuki Izumi and Ai Tateishi at Sony Group Corporation for software integration. This research was partially supported by the ETH AI Center. Thanks to Takahiro Miki, Yuntao Ma, Kento Kawaharazuka at ETH Zurich for reinforcement learning discussion.

\clearpage

\bibliographystyle{IEEEtran}
\bibliography{main}

\end{document}